\documentclass[sigconf]{acmart}

\usepackage{enumitem}
\usepackage{mathrsfs}
\usepackage{algorithm}
\usepackage{algorithmic}
\usepackage{amsmath}
\usepackage{balance}
\AtBeginDocument{%
  \providecommand\BibTeX{{%
    \normalfont B\kern-0.5em{\scshape i\kern-0.25em b}\kern-0.8em\TeX}}}

\copyrightyear{2020} 
\acmYear{2020} 
\setcopyright{acmcopyright}
\acmConference[MM '20]{Proceedings of the 28th ACM International Conference on Multimedia}{October 12--16, 2020}{Seattle, WA, USA}
\acmBooktitle{Proceedings of the 28th ACM International Conference on Multimedia (MM '20), October 12--16, 2020, Seattle, WA, USA}
\acmPrice{15.00}
\acmDOI{10.1145/3394171.3413777}
\acmISBN{978-1-4503-7988-5/20/10}
\usepackage{enumitem}

\acmSubmissionID{1988}

\begin{document}
\fancyhead{}
\title{IR-GAN: Image Manipulation with Linguistic Instruction \\ by Increment Reasoning}
\renewcommand{\shortauthors}{}
%\renewcommand\footnotetextcopyrightpermission[1]{} 
\author{Zhenhuan Liu$^{1,2}$, Jincan Deng$^{1, 2}$, Liang Li$^{1, *}$, Shaofei Cai$^{1,2}$, Qianqian Xu$^{1}$, Shuhui Wang$^{1}$, Qingming Huang$^{1,2,3}$}
\thanks{$^\ast$Corresponding author.}
\affiliation{%
    \institution{$^1$Key Laboratory of Intelligent Information Processing, Institute of Computing Technology, CAS, Beijing, China}
    %\institution{$^2$University of Chinese Academy of Sciences, Beijing, China}
    \institution{$^2$School of Computer Science and Technology, University of Chinese Academy of Sciences, Beijing, China}
    \institution{$^3$Peng Cheng Laboratory, Shenzhen, China}
    %\institution{$^3$Key Laboratory of Big Data Mining and Knowledge Management, Chinese Academy of Sciences, Beijing, China}
    % \institution{$^3$School of Information Science and Technology, University of Science and Technology of China, Hefei, China}
}
\email{{zhenhuan.liu, jincan.deng, shaofei.cai}@vipl.ict.ac.cn, {liang.li, xuqianqian, wangshuhui}@ict.ac.cn, qmhuang@ucas.ac.cn}

\begin{abstract}
    Conditional image generation is an active research topic including text2image and image translation.
    Recently image manipulation with linguistic instruction brings new challenges of multimodal conditional generation.
    However, traditional conditional image generation models mainly focus on generating high-quality and visually realistic images, and lack resolving the partial consistency between image and instruction.
    To address this issue, we propose an Increment Reasoning Generative Adversarial Network (IR-GAN), which aims to reason the consistency between visual increment in images and semantic increment in instructions.
    First, we introduce the word-level and instruction-level instruction encoders to learn user's intention from history-correlated instructions as semantic increment.
    Second, we embed the representation of semantic increment into that of source image for generating target image, where source image plays the role of referring auxiliary.
    Finally, we propose a reasoning discriminator to measure the consistency between visual increment and semantic increment, which purifies user's intention and guarantees the good logic of generated target image.
    Extensive experiments and visualization conducted on two datasets show the effectiveness of IR-GAN.
\end{abstract}



\maketitle
\vspace{-0.2cm}
{
    \fontsize{8pt}{8pt}
    \selectfont
    \textbf{ACM Reference Format:}\\
    Zhenhuan Liu, Jincan Deng, Liang Li, Shaofei Cai, Qianqian Xu, Shuhui Wang, Qingming Huang. 2020. IR-GAN: Image Manipulation with Linguistic Instruction by Increment Reasoning. In \textit{Proceedings of the 28th ACM International Conference on Multimedia (MM’20), October 12--16, 2020, Seattle, WA, USA.} ACM, New York, NY, USA, 9 pages. https://doi.org/10.1145/3394171.3413777
}
\vspace{-0.2cm}
\section{Introduction}
\begin{figure}[t]
    \centering
    \includegraphics[width=\linewidth]{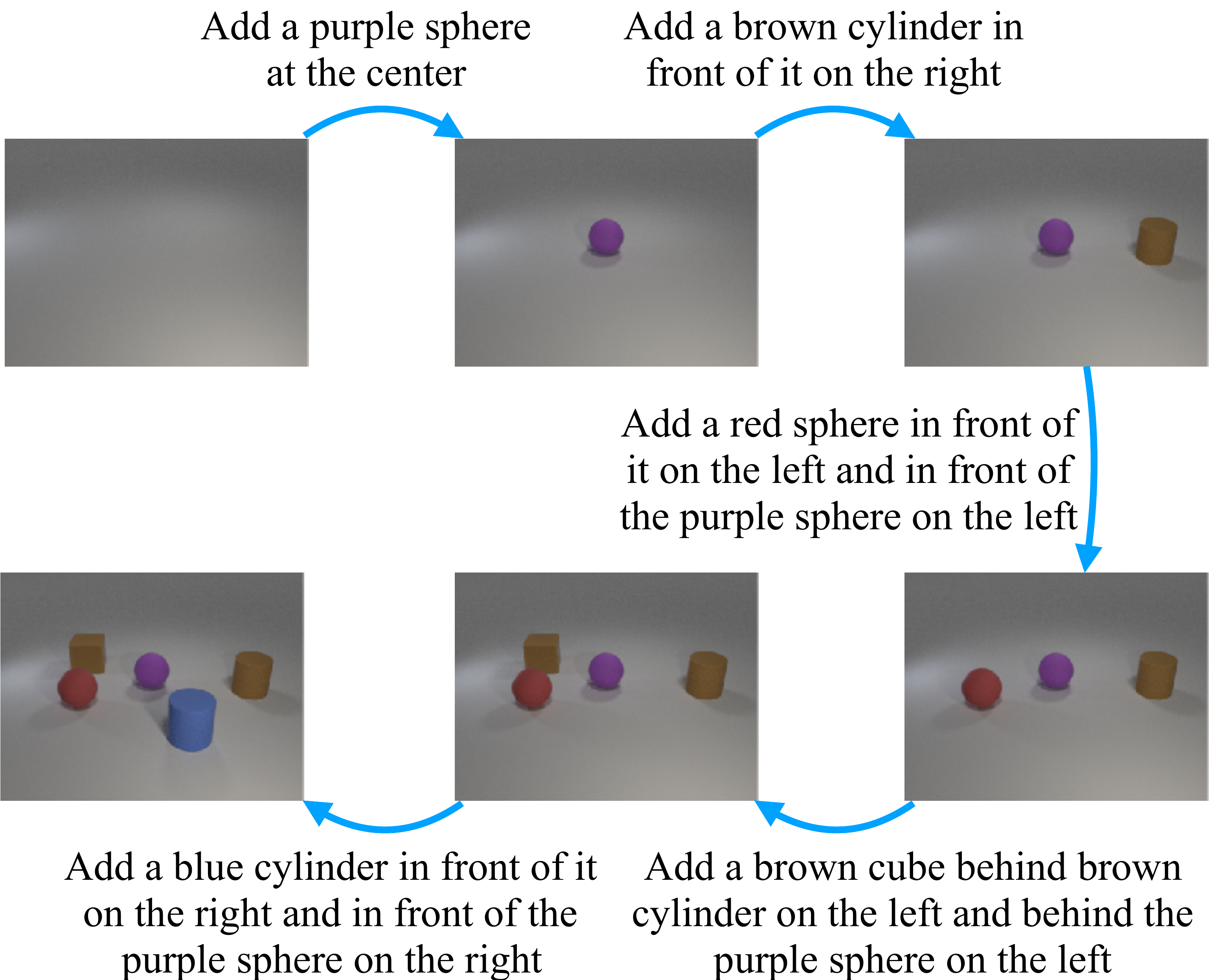}
    \caption{Overview of image manipulation with linguistic instruction task, at each time step, the source image is manipulated into a new image according to current linguistic instruction.}
    \Description{task of this work}
    \label{fig:task}
    % \vspace{-0.8cm}
\end{figure}

% introduction
Conditional image generation allows people produce preferred
image based on their intention. Recent research have made this with different types of conditions,
such as class label, text description or source images for image translation.
As a new conditional generation task, image manipulation with linguistic instruction has attracted a lot of attention.
Figure~\ref{fig:task} shows the flowchart of this task: given an image and a series of sequential linguistic instructions,
this task is to iteratively manipulate the image by understanding user's intention.
This brings a more natural and effective human-AI interaction,
and has great significance in real applications, such as image processing, content creation, etc.

% Task analysis
As a multimodal conditional image generation task, image manipulation
with linguistic instruction is more complex than other unimodal conditional generation tasks.
There exist three key issues in this task:
first, perceive the visual content from source image, and cognize the user's intention from linguistic instruction;
second, align the referring phrase in the instruction with the visual elements in the source image,
and reason the location, appearance and attributes of new changes for target image;
third, construct the image generation model based on multimodal inputs.
For current generation technologies, these issues are still big challenges to be solved.

Conditional image generation tasks can be divided into five groups by their condition types.
% traditional image generation
(1) Label-based conditional image generation~\cite{mirza2014conditional, odena2017conditional,yan2016attribute2image,marek2020config, yi2020attentive, liu2019adaptive, li2012learning}.
It uses class or visual attributes as labels to generate realistic images.
Due to the simplicity of condition, we can hardly control the details of generated images.
% Image-to-image generation
(2) Image-to-Image(I2I) generation~\cite{isola2017image,zhu2017unpaired,park2019semantic,lee2020drit++,lin2020multimodal, yang2019skeletonnet, yang2017graph}.
This task is to translate source image in a domain to the style of another domain.
I2I methods can generate natural and realistic images but it requires that the images in two domains share some common geometry structures.
% Text to image generation
(3) Text-to-Image(T2I) generation~\cite{reed2016generative,zhang2018stackgan++,xu2018attngan,tan2019semantics,gao2019perceptual,sharma2018chatpainter, zhang2020state}.
Its goal is to generate images according to a linguistic sentence.
In the early stage, researchers introduced visually-discriminative vector representation of text descriptions as condition to generate images.
Recently, some works use different architectures to generate high resolution images and strengthen the semantic consistency with text description.
Due to the limited amount of information in a sentence, T2I is hard to generate satisfying images for users.
% Semantic image synthesis
(4) Semantic image synthesis(SIS)~\cite{dong2017semantic,nam2018text}.
Dong \textit{et al.}~\cite{dong2017semantic} first tried to implement such task,
which changes the source image to match with the text description.
Although this task can do some manipulation on the origin image,
it only focuses on changing the attributes of different objects while keeping the global structure.
% image manipulation
(5) Image Manipulation with linguistic instruction(\textbf{IM})~\cite{el2019tell}.
This task iteratively generates images under multimodal condition including a source image and a sequence of text instructions.
El-Nouby \textit{et al.}~\cite{el2019tell} proposed a recurrent image generation model
which takes into account both current source image and past text instructions.
By representing intention by multi-step instructions and getting dynamic feedback of manipulated image,
user can guide the generation process more flexibly and accurately.

% challenge description
In the IM task, it requires not only perceiving the content of each condition,
but also cognizing user's manipulation intention from two conditions.
However, the elements in both conditions are only partially consistent:
(1) The referring phrase in instructions are consistent with existing visual elements in images.
(2) The semantic information about new objects in instructions are not represented in current image.
(3) The image includes background, prior knowledge of objects appearance that may be irrelevant with the instructions.
This cross-modal partial consistency brings difficulty to the generation of target image.

% Method
In this paper, we propose the Increment Reasoning Generative Adversarial Network (IR-GAN) to model the multimodal conditional generation for image manipulation.
The increment reasoning mechanism is designed to reason the consistency between visual increment in target image and user's intention in linguistic instruction.
Our model consists of instruction encoder, image generator and reasoning discriminator.
%(1)instruction encoder
For the instruction encoder, we use word-level and instruction-level GRUs as encoder to
learn user's intention from current instruction along with history instruction informations.
The user's intention is purified into the increment representation through the backpropagation of the discriminator.
%(2)generator
For the generator, we introduce a MLP to project the above representation to a semantic increment feature map,
and embed it into the feature map of source image.
Then the composited features are fed into the image decoder to generate target image,
where source image plays the role of referring auxiliary.
%(3)discriminator
Finally, we propose a reasoning discriminator to reason the consistency among
existing visual elements, visual increment and the corresponding instruction.
This multimodal conditional discriminator can guarantee the good logic of generated target image.
%(4)summary
By explicitly modeling the visual increment in image and the semantic increment in instructions,
we formulate the interactions between image and text in the generation of target image.

% contribution
The contributions of this paper are summarized as follows:
\begin{enumerate}
    \item We propose the IR-GAN to model the task of image manipulation with linguistic instruction,
          which learns user's intention from instructions and iteratively manipulates images.
          \vspace{0.1cm}
    \item We design an adversarial mechanism of increment reasoning, where the generator is to generate the visual increment and the discriminator is to figure out its consistency with semantic increment.
          \vspace{0.1cm}
    \item We evaluate our model on two datasets. Extensive experiments show that our model
          surpasses the state-of-the-art performance.
\end{enumerate}

\section{Related Work}
\textbf{Image-to-Image Translation(I2I).} Originally many works focused on different tasks of image-to-image translation,
such as image inpainting, image super-resolution, style transfer and each one employs a specific loss formulation.
Isola \textit{et al.} \cite{isola2017image} proposed pix2pix framework to implement a general purpose image-to-image translation architecture,
which applies adversarial loss to learn the translation with paired image.
Then Zhu \textit{et al.} \cite{zhu2017unpaired} proposed CycleGAN to resolve
the unpaired image-to-image translation problem with an additional cycle loss.
Recently Park \textit{et al.} \cite{park2019semantic} came up with  spatially-adaptive normalization
to generate realistic image from a semantic segmentation mask.

\textbf{Text-to-Image Generation(T2I).} The goal of T2I is to generate realistic image which captures the representation given by the text description.
Reed \textit{et al.} \cite{reed2016generative} first employed conditional generative adversarial networks (cGANs) to implement T2I and verified its effectiveness.
To generate high resolution images, Zhang \textit{et al.} \cite{zhang2018stackgan++} leveraged multiple generators and discriminators to model the image in different scales.
Then Xu \textit{et al.} \cite{xu2018attngan} applied attention mechanism to focus on different semantic features at each subregion of images
during different stages of image generation.
Recently, Qiao \textit{et al.} \cite{qiao2019mirrorgan} designed an additional I2T loss to enhance the semantic consistency between generated image and source text by redescription.
Due to the gap between image and text modality,
learning the appropriate image representation of text is still a challenge to be solved.

\begin{figure*}[t]
    \centering
    \includegraphics[width=\linewidth]{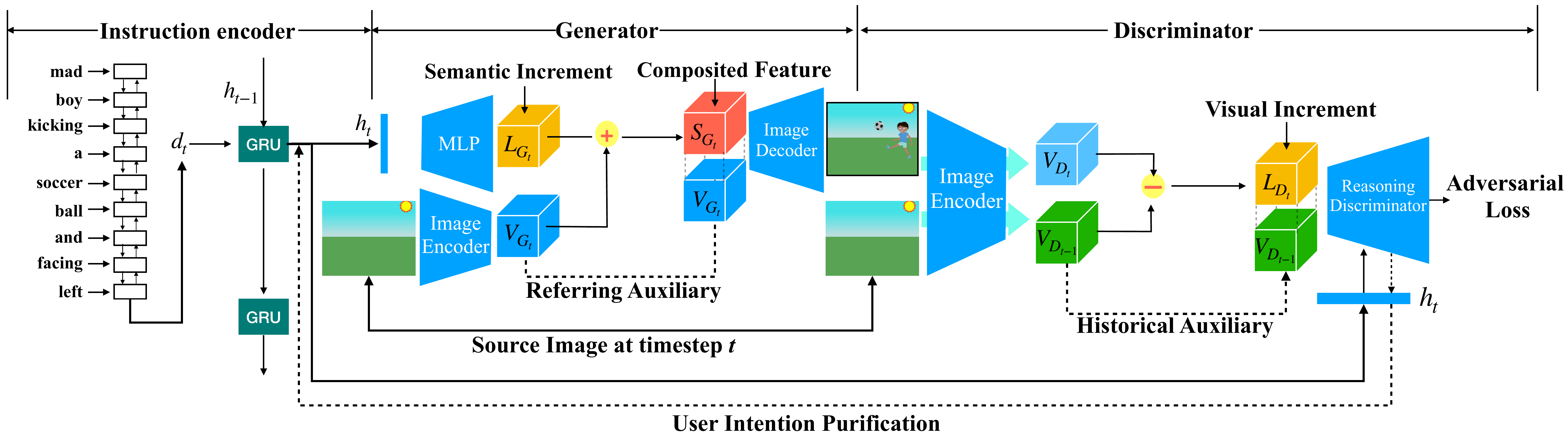}
    \caption{Overview of IR-GAN structure.
        At each time step ${t}$, the instruction encoder learns user's intention representation $h_t$
        through current instruction representation $d_t$ and history instruction information.
        The generator first projects $h_t$ to visual feature space and get the feature map $L_{G_{t}}$ representing semantic increment.
        After that, we composite $L_{G_{t}}$ with $I_{G_t}$ by addition and get $S_{G_t}$ as an abstract representation of target image,
        which then is fed into the image decoder to generate target image.
        For the discriminator, we first extract the visual increment feature $L_{D_t}$ by subtraction between the two feature maps, $I_{D_t}$ and $I_{D_{t-1}}$.
        And then we leverage a multimodal discriminator to determine its consistency with $h_t$.
        It's worth noting that the instruction encoder is only optimized through the backpropogation from discriminator, which ensures
        the learned instruction representation semantic $h_t$ are consistent with visual increment.
    }
    \Description{sd}
    \label{fig:structure}
\end{figure*}

\textbf{Semantic Image Synthesis(SIS).} Different from I2I and T2I, semantic image synthesis is a multimodal conditional image generation task conditioned on both image and text.
It aims to generate realistic images not only matching the target text description, but also maintaining other image features that are irrelevant to the text description.
Dong \textit{et al.} \cite{dong2017semantic} first proposed an end-to-end neural architecture built upon cGANs framework.
Nam \textit{et al.} \cite{nam2018text} designed a text-adaptive GANs model which uses
word-level local discriminators to classify fine-grained attributes of images independently.
However, the main purpose of SIS is for text-to-image synthesis, rather than using text to manipulate image semantically.

\textbf{Image Manipulation with linguistic instruction(IM).}
This task aims to manipulate visual elements in the image following linguistic instructions, such as adding new objects, changing layout and attributes of objects.
El-Nouby \textit{et al.} \cite{el2019tell} first implemented this image manipulation task with sequential linguistic instructions.
They proposed a recurrent architecture to generate the final image through a sequence of image manipulation process, which demonstrates much better performance than generating final image in a single pass.
Due to the complex interactions between image and linguistic instruction, this task needs complete and fine-grained understanding of multimodal inputs, thoroughly reasoning of user's intention, and accurately generating new changes for target image.

\section{Method}
\subsection{Overview}
To address the IM task, we propose the Increment Reasoning Generative Adversarial Network, which consists
of instruction encoder, image generator and reasoning discriminator.
The whole architecture is shown in Figure~\ref{fig:structure}.
As the task involves a sequence of linguistic instructions, we propose iterative algorithm to
learn user's intention from instructions and manipulate the image correspondingly.
In each iterative process of manipulation, we explore the historical instructions as auxiliary information,
instead of only relying on current instruction.
First, The instruction encoder is designed to embed the instruction information into feature vectors.
Second, guided by user's intention representation from instruction encoder,
the image generator manipulates source image into target image.
Finally, we leverage the reasoning discriminator to measure the consistency among existing visual elements, visual increment and corresponding instruction,
which directs the image generator to generate target images with good logic.

\subsection{Instruction Encoder}
Since instructions at different time steps are correlated, the current instruction can not represent user's intention alone.
As illustrated in Figure~\ref{fig:task}, all instructions except for the first one use "\textit{it}" to indicate the object added at last time step,
instead of directly referring to the visual element in source image.
To resolve such problem, we propose the hierarchical instruction encoder
to learn the representation of user's intention by consuming the current instruction along with history instruction information,
which consists of a word-level encoder and an instruction-level encoder.

Specifically, let $\{s_1, s_2...s_n\}$ denote the sequence of instructions and $\{w_{t}^1, w_{t}^2...w_{t}^m\}$ denote the embeddings of each word in $s_t$.
For the word-level encoder, we utilize a bidirectional GRU to encode the current instruction $s_t$ and compute the last hidden state $d_t$ as the representation of instruction.
For the instruction level encoder, we utilize another GRU to learn representation $h_t$ of history instructions.
The above process can be formulated as,
\begin{align}
    d_t & = BiGRU(w_{t}^1, w_{t}^2...w_{t}^m) \\
    h_t & = GRU(d_t, h_{t-1})
\end{align}
where $w_{t}^m \in \mathbb{R}^K$, $d_t \in \mathbb{R}^N$ and $h_t, h_{t-1} \in \mathbb{R}^M$,
$K,N,M$ denote the dimension of word embeddings, word-level encoder hidden state, and instruction-level encoder hidden state, respectively.
Besides, we introduce the conditional augmentation~\cite{zhang2018stackgan++} to augment the feature vector of linguistic instruction,
which helps generate more descriptive feature vectors and makes model more robust to new instructions.
The instruction encoder is only optimized through the backpropagation of reasoning discriminator,
thus it is encouraged to purify the historical instruction information into
increment representation of user's intention consistent with visual increment at this time step.

\subsection{Image Generator}
% summary
Guided by the learned feature vector from instruction encoder,
we use the image generator to manipulate the source image into target image, which can be formulated as,
\begin{equation}
    \tilde{I_t} = G(I_{t-1}, h_t)
\end{equation}
where $I_{t-1}$, $\tilde{I_t}$ denote the source image and the generated target image at this time step, and $h_t$ denotes the instruction representation.
The image manipulation process can be decomposed into three steps:
(1) perceive the visual content from source image, the semantic increment from the instruction.
(2) embed the semantic increment on the source image.
(3) generate target image based on the composited feature. We design the three modules of image generator, respectively.

\textbf{Image Encoder.}
To perceive the visual information from source image, we utilize a shallow Convolution Neural Network(CNN) as $E_G$ to encode the source image $I_{t-1}$ into the feature map $V_{G_{t-1}} \in \mathbb{R}^{C\times H\times W}$, which can be formulated as,
\begin{equation}
    V_{G_{t-1}} = E_G(I_{t-1})
\end{equation}
Specifically, the image encoder consists of 4 residual building blocks (ResBlocks) with kernel size of $3\times3$ and Average Pooling stride of $2\times2$.
Finally, we obtain the feature map $V_{G_{t-1}}$ of source image with the spatial size of $16\times 16$.

\textbf{Embedding of Semantic Increment.}
% One of the key issues in this task is to cognize user's intention from the representation of source image and linguistic instruction.
One of the key issues in this task is to aggregate semantic information and visual representation to generate target image.
To solve this, we project the purified semantic increment representation $h_t$
to the representation $L_{G_t}$ in visual space using a Multiple Layer Perception(MLP).
Then, we use element-wise addition between $V_{G_{t-1}}$ and $L_{G_t}$ to generate an composited multimodal representation $S_{G_t}$,
which can be formulated as,
\begin{equation}
    S_{G_t} = V_{G_{t-1}} + L_{G_t}
\end{equation}
The semantic increment $L_{G_t}$ represents the changed manipulation information of instruction.
% Informally, it seems like a composition process in which a new image is layered on top of the base image.
By using addition operation in the spatial dimension,
the referring semantic information in instruction will be aligned with the corresponding visual representation of source image,
and the semantic increment information will generate the visual increment of target image.

\textbf{Image Decoder.}
In this module, we utilize stacked transposed convolution layers to synthesize target image based on the composited representation.
However, it is not enough to rely on $S_{G_t}$ alone to generate the target image, since it doesn't cover all detailed visual information of source image.
To solve the problem, we introduce $V_{G_{t-1}}$ as auxiliary information to help the decoder construct the target image.
\begin{equation}
    \tilde{I}_t = F_G([S_{G_t};V_{G_{t-1}}])
\end{equation}

Following the work~\cite{miyato2018cgans}, we apply conditional batch normalization and spectral normalization~\cite{miyato2018spectral} at each layer of the image decoder.

\subsection{Reasoning Discriminator}
To measure the quality and logic of generated target image, we propose the reasoning discriminator,
which is to reason the consistency between visual increment upon source image and user's intention in linguistic instruction.
This discrimination process can be formulated as,
\begin{equation}
    r = D(\tilde{I}_t, I_{t-1}, h_t)
\end{equation}
Where $r\in R$ is the output score, $\tilde{I}_t$ is the generated target image, $I_{t-1}$ is the source image.
After obtaining the visual information of both source and target image,
we extract the visual increment between them and use a multimodal discriminator to determine it's consistency with user's intention.

\textbf{Image Encoder.} Similar with the above image encoder of generator, we use another CNN to encode both source image and target image,
and compute their feature maps of $V_{G_{t-1}}$ and $V_{G_t}$.
By projecting source and target image into a common feature space,
we implement the alignment of their visual information.

\textbf{Extraction of Visual Increment.}
In a single image manipulation stage, most visual elements in both source image and target image are common, and they are irrelevant with current user's intention.
The semantic difference between two images is the key to determine the logic of the generated target image.
We employ element-wise subtraction between feature maps of two images to extract the difference
and obtain the visual increment $L_{D_t}$, which can be formulated as,
\begin{equation}
    L_{D_t} = V_{D_t} - V_{D_{t-1}}
\end{equation}
where $L_{D_t} \in \mathbb{R}^{C \times H \times W}$ can be regarded as the feature representation of the visual changes
in the discrimination process.

\textbf{Multimodal Discriminator.}
Finally, we leverage a multimodal discriminator to measure the quality and logic of generated target image under the given instructions.
Since the error of generated image could be caused by history wrong manipulation, relying on visual increment alone is not enough.
Here we employ the source image feature map as an auxiliary information.
In detail, following the cGANs~\cite{miyato2018cgans}, we use a projection discriminator to measure the consistency of this multimodal condition.
The projection discriminator can be formulated as,
\begin{eqnarray}
    x_t = \phi([L_{D_t}; V_{D_{t-1}}]) \\
    r = h_t^Tx_t + \psi(x_t)
\end{eqnarray}
where $V_{D_{t-1}}$ is the feature map of source image, $\phi$ is a MLP module to project visual feature maps to a feature vector with the same dimension as $h_t$,
$\psi$ is a fully connected layer to project $x_t$ into a scalar.

\subsection{Objective Optimization}
In the conditional adversarial learning procedure,
the discriminator is to distinguish the generated fake image under given condition, while the generator tries to fool the discriminator.
Following the common practice, the loss function is designed for two purposes.
One is to ensure the generated image visually realistic,
and the other is to make it consistent with corresponding instruction.
In the multimodal conditional case, the discriminator may focus too much on visual details
and ignore the global influence of linguistic condition.
Thus, besides the loss about real and fake, we introduce the loss about inconsistent pairs to encourage the discriminator to formulate the consistence between visual increment and user's intention.
In IR-GAN, the objective of discriminator is to minimize this hinge loss:
\begin{equation}
    L_{D}=L_{D_{\mathrm{real}}}+\alpha L_{D_{\mathrm{fake}}}+ \beta L_{D_{\mathrm{inconsistence}}}
\end{equation}
Where
\begin{align}
    L_{D_{\mathrm{real}}}   =-\mathbb{E}_{(I_{t}, I_{t-1}, h_{t}) \sim p_{\mathrm{data}(0: T)}}[\min (0,-1+D(I_{t}, I_{t-1}, h_t))]   \\
    L_{D_{\mathrm{fake}}}   =-\mathbb{E}_{(I_{t-1}, h_t) \sim p_{\mathrm{data}(0: T)}}[\min (0,-1-D(G(I_{t-1}, h_t), I_{t-1}, h_t ))] \\
    L_{D_{\mathrm{inconsistence}}}  =-\mathbb{E}_{(I_{t}, I_{t-1}, \tilde{h_t}) \sim p_{\mathrm{data}(0: T)}}[\min (0,-1-D(I_{t}, I_{t-1}, \tilde{h_t} ))]
\end{align}
$\tilde{h_t}$ is the instruction feature vector under the condition of inconsistent pairing,
$\alpha$ and $\beta$ are the loss weight to handle the importance about fake loss and the loss of inconsistent pairs.
% In the optimization process, the discriminator are directed to give positive score for the true condition pair,
% and give negative score for the generated condition pair and the wrong condition pair.
During the adversarial training of IR-GAN,
the objective loss directs the discriminator to measure the consistency between visual increment in images and semantic increment in instructions.
In the meantime, the instruction encoder is optimized using the backpropagation gradient from the discriminator,
which encourages it to purify the user's intention into the increment information consistent with the visual increment in the manipulation process.

The objective of generator is to minimize this loss based on discriminator:
\begin{equation}
    L_{G} = -\mathbb{E}_{(I_{t-1}, h_t) \sim p_{\mathrm{data}(0: T)}}D(G(I_{t-1}, h_t))
\end{equation}
This loss encourages generator to composite the learned increment representation from instruction encoder
with source image to fool the discriminator,
which ensures the quality and good logic of the generated target image.

\section{Experiments}

\subsection{Datasets}
For this image manipulation task with a sequence of instructions,
the standard dataset contains linguistic instructions describing manipulation actions and
corresponding ground truth images for each instruction.
Here we use two standard benchmarks, namely CoDraw~\cite{kim2019codraw} and i-CLEVR~\cite{el2019tell}.
% Examples of the two datasets are shown in Figure~\ref{fig:task} and Figure~\ref{fig:codraw}.

\textbf{CoDraw}~\cite{kim2019codraw} is a dataset for collaborative drawing between a teller and a drawer.
Firstly the teller is provided a final target image, and at each turn the teller gives a text instruction to the drawer.
Based on the drawer's generated image, the teller gives the next instruction to drawer until the generated image is similar enough to the final target image.
This dataset consists of 9993 samples totally, and each sample consists of varying length conversations.
The target image in each sample contains different objects of 58 types, such as trees, planes and balls.
And each object has the different size, direction and location.

\textbf{i-CLEVR} CLEVR~\cite{johnson2017clevr} is a programmatically rendered dataset and often used in Visual Question Answering(VQA) tasks.
Each image in CLEVR consists of several objects with different shapes, colors and sizes. Based on the generation code of CLEVR, i-CLEVR
generates a sequence of (image, instruction) pairs starting from an empty canvas. Each instruction describes a new object to be added with
its shape and color. The place of the new added object is specified by trelative location to existing objects. This dataset contains
10,000 sequences, and each sequence contains 5 linguistic instructions.

\subsection{Evaluation Metrics}
For GAN based models, the Inception Score(IS)\cite{salimans2016improved} or
Fr$\acute{\text{e}}$chet Inception Distance (FID)\cite{heusel2017gans} are popularly used
to measure the diversity and reality relative to the true image distribution. In this image manipulation
task, the correctness of the manipulation result according to user's intention is
the most important evaluation.
For a good generated result, the objects mentioned in user's instruction should not only be generated accurately,
but also follow the layout indicated by user's intention.

To meet these constraints, we use object localizer pretrained on the training dataset
to detect the objects and its location of both ground true image and generated image.
For each generated example, we compare the detected results and compute the precision, recall and F1 score.
Besides, to measure the arrangement of generated objects, we compute the relational similarity score
as~\cite{el2019tell} by constructing a scene graph for each example, in which the objects and image center point are the vertices and their left-right,
front-back relations are the directed edges. The relational similarity is computed by the percentage of common relations between generated image and ground true image:
\begin{equation}
    \operatorname{rsim}\left(E_{G_{\mathrm{gt}}}, E_{G_{\mathrm{gen}}}\right)=\operatorname{recall} \times \frac{\left|E_{G_{\mathrm{gen}}} \cap E_{G_{\mathrm{gt}}}\right|}{\left|E_{G_{\mathrm{gt}}}\right|}
\end{equation}
Where "recall" is the recall over detected objects of the generated image to that in the ground truth image.
$E_{G_{\mathrm{gt}}}$ is the set of relational edges for the ground truth image corresponding to vertices common to both ground truth images and generated images
and $E_{G_{\mathrm{gen}}}$ is the set of relational edges for the generated image corresponding to vertices common to both ground truth images and generated images.

\subsection{Implementation Details}
% model
We use GloVe\cite{pennington2014glove} as the input word embeddings of the word-level instruction encoder.
The dimension of hidden state in word-level encoder and instruction-level encoder are both set as 1024.
The image encoder of generator and discriminator both use ResBlocks with $3\times3$ kernel and $2\times2$ Average Pooling.
The dimension of $V_{G_t}$, $V_{D_t}$ and $L_{G_t}$ are $256\times 16\times 16$.

% regularization
We add the layer normalization\cite{ba2016layer} in the word-level encoder and instruction-level encoder.
We introduce batch normalization\cite{ioffe2015batch} in each layer of the image encoder of generator and spectral normalization~\cite{miyato2018spectral}
for all layers of the discriminator.

% training
During training\footnote{The whole source code will be released on github.}, we use the ground truth image $I_{t-1}$ as input to generate $\tilde{I}_{t}$, but use $\tilde{I}_{t-1}$ for the test time.
We use Adam\cite{kingma2015adam} to optimize the parameters of generator and discriminator with momentums of 0 and 0.9, and instruction encoder
of with momentums of 0.999 and 0.9.
The learning rate of generator, discriminator, word-level encoder, instruction-level encoder are set to 0.0001, 0.0004, 0.003 and 0.006 respectively.
For the training dynamics, the generator and discriminator parameters are updated at every time step,
while the parameters of instruction encoder are updated in every sequence.
The instruction encoder is trained with respect to the discriminator objective only.

\subsection{Experimental Results}

\begin{figure*}[h]
    \centering
    \includegraphics[width=\linewidth]{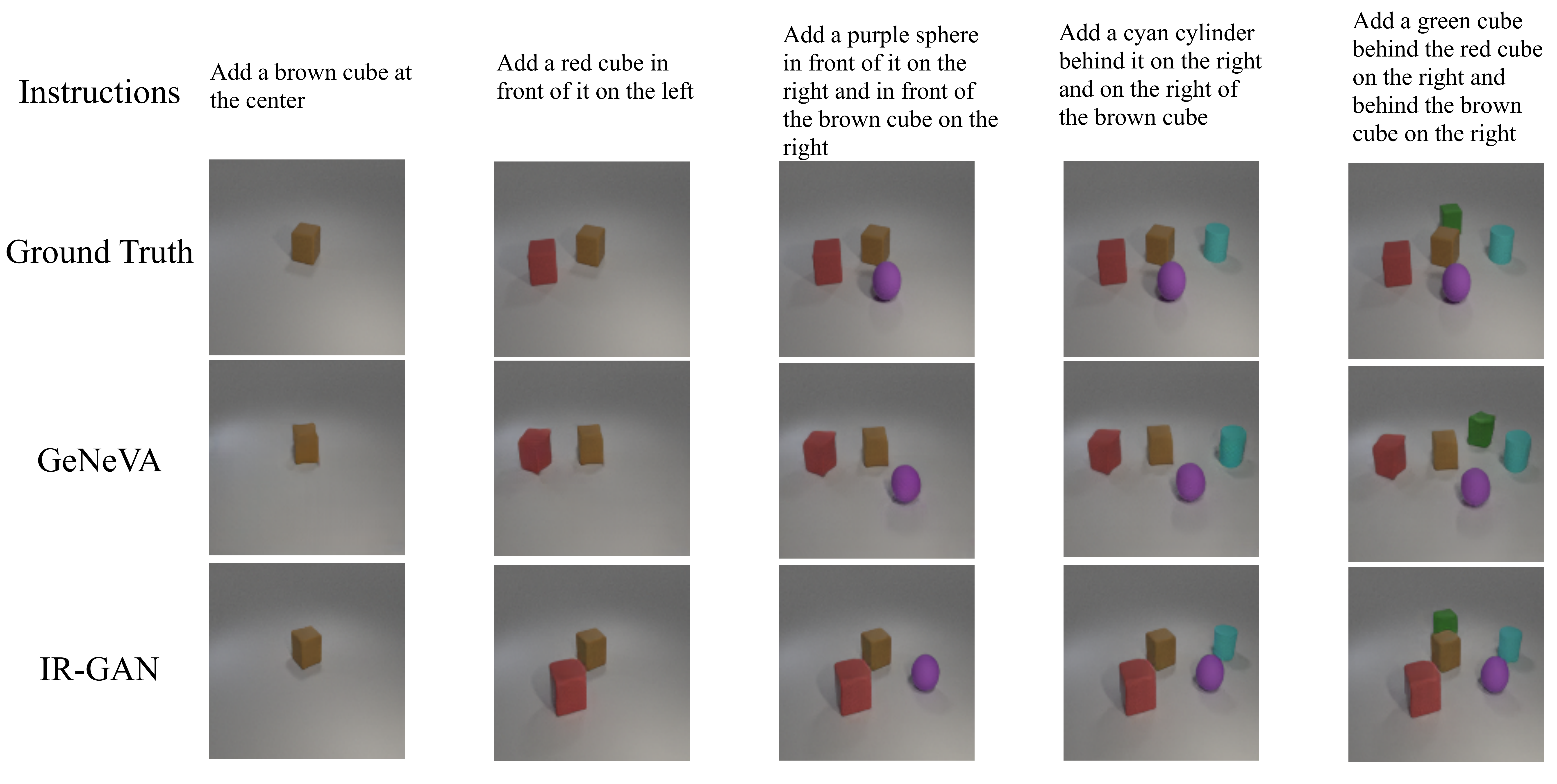}
    \caption{Examples of image manipulation with linguistic instruction on i-CLEVR dataset.
        The first row represents the instruction sequence in a complete manipulation process.
        The second row shows the ground truth image.
        The third and fourth row are the generated images by state-of-the-art GeNeVA~\cite{el2019tell} and our IR-GAN.
        The resolution of images are scaled to 128x128 in the pre-processing step.}
    \label{fig:iclevr_example}
    \Description{..}
\end{figure*}

To evaluate the effectiveness of the proposed IR-GAN, we conduct experiments on the i-CLEVR and CoDraw dataset
and compare the performance with several state-of-the-art models.
The single-step text2image model leverages general text2image architecture and only uses
the concatenation of sequential instructions as the condition to generate a single final image.
The iterative text2image model iteratively generates images based on history instruction.
The iterative IM model uses both image and instructions as the condition to generate the target image.
GeNeVA~\cite{el2019tell} is current state-of-the-art model for this task.
To ensure the equity of comparison, we use the same pretrained detection models as GeNeVA to detect the elements in generated target images, and the
split of training and testing dataset are also identical.

\subsubsection{Results on i-CLEVR dataset}
Table~\ref{tab:clevr} shows the performance of different methods.
We can observe that, first, our IR-GAN surpasses the state-of-the-art method GeNeVA~\cite{el2019tell} on all four metrics.
By introducing increment reasoning, IR-GAN can focus on the increment semantic intention in the instruction encoding process
and increment visual elements in the image generation process, which better formulates the interactions between image and instruction.
Second, the high precision score reflects the ability of our model to generate visual elements from semantic representation in the image;
the high recall score indicates our model can thoroughly capture the semantic meaning in the instruction;
the relational similarity score proves IR-GAN is good at understanding and reasoning the interaction between image and linguistic instruction.
Third, the F1 score of our method achieves the 90\%+ performance, which means it has the potential for practical applications.

\begin{table}[t]
    \centering
    \caption{Results comparison of image manipulation with linguistic instruction on i-CLEVR Dataset}
    \label{tab:clevr}
    \begin{tabular}{|l|l|l|l|l|}
        \hline
        Method                 & Precision      & Recall         & F1             & Rsim           \\
        \hline
        Single-step text2image & 25.49          & 20.95          & 22.63          & 11.52          \\
        Iterative text2image   & 71.15          & 60.57          & 65.44          & 50.21          \\
        Iterative IM           & 88.47          & 83.35          & 85.83          & 70.22          \\
        GeNeVA                 & 92.39          & 84.72          & 88.39          & 74.02          \\
        \hline
        Our IR-GAN             & \textbf{94.81} & \textbf{86.90} & \textbf{90.68} & \textbf{74.26} \\
        \hline
    \end{tabular}
    % \vspace{-0.5cm}
\end{table}

\subsubsection{Results on CoDraw dataset}
The performance comparison on CoDraw dataset is shown in Table~\ref{tab:codraw}, and our model surpasses state-of-the-art for all metrics.
First, we find that similar with the experiments on i-CLEVR, our model surpasses all other methods on precision, recall, F1 score and relational similarity.
This proves the effectiveness of increment reasoning again in this multimodal conditional image generation task.
Second, we notice that compared with i-CLEVR, the performance of our IR-GAN on this dataset is relatively low.
The reason lies in that
(1) the visual elements in CoDraw are much more complex and various than those in i-CLEVR dataset;
(2) different from the instruction in i-CLEVR generated programmatically via predefined template, the instructions in CoDraw are generated
by human interaction. The teller uses the detailed information like the horizon line in the background,
which needs the models have the stronger cognition.
(3) the conversations between teller and drawer are concatenated together until the image is changed, which can cause the longer term of
instructions. However, the instruction encoder is hard to capture information for very long instructions.

\begin{table}[t]
    \centering
    \caption{Results comparison of image manipulation with linguistic instruction on CoDraw Dataset}
    \label{tab:codraw}
    \begin{tabular}{|l|l|l|l|l|}
        \hline
        Method                 & Precision      & Recall         & F1             & Rsim           \\
        \hline
        Single-step text2image & 50.60          & 43.42          & 44.96          & 22.33          \\
        Iterative text2image   & 62.47          & 48.95          & 54.89          & 32.74          \\
        Iterative IM           & 66.38          & 51.27          & 57.85          & 33.57          \\
        GeNeVA                 & 66.64          & 52.66          & 58.83          & 35.41          \\
        \hline
        Our IR-GAN             & \textbf{70.12} & \textbf{52.75} & \textbf{60.20} & \textbf{35.51} \\
        \hline
    \end{tabular}
    \vspace{-0.5cm}
\end{table}

\begin{figure*}[t]
    \centering
    \includegraphics[width=\linewidth]{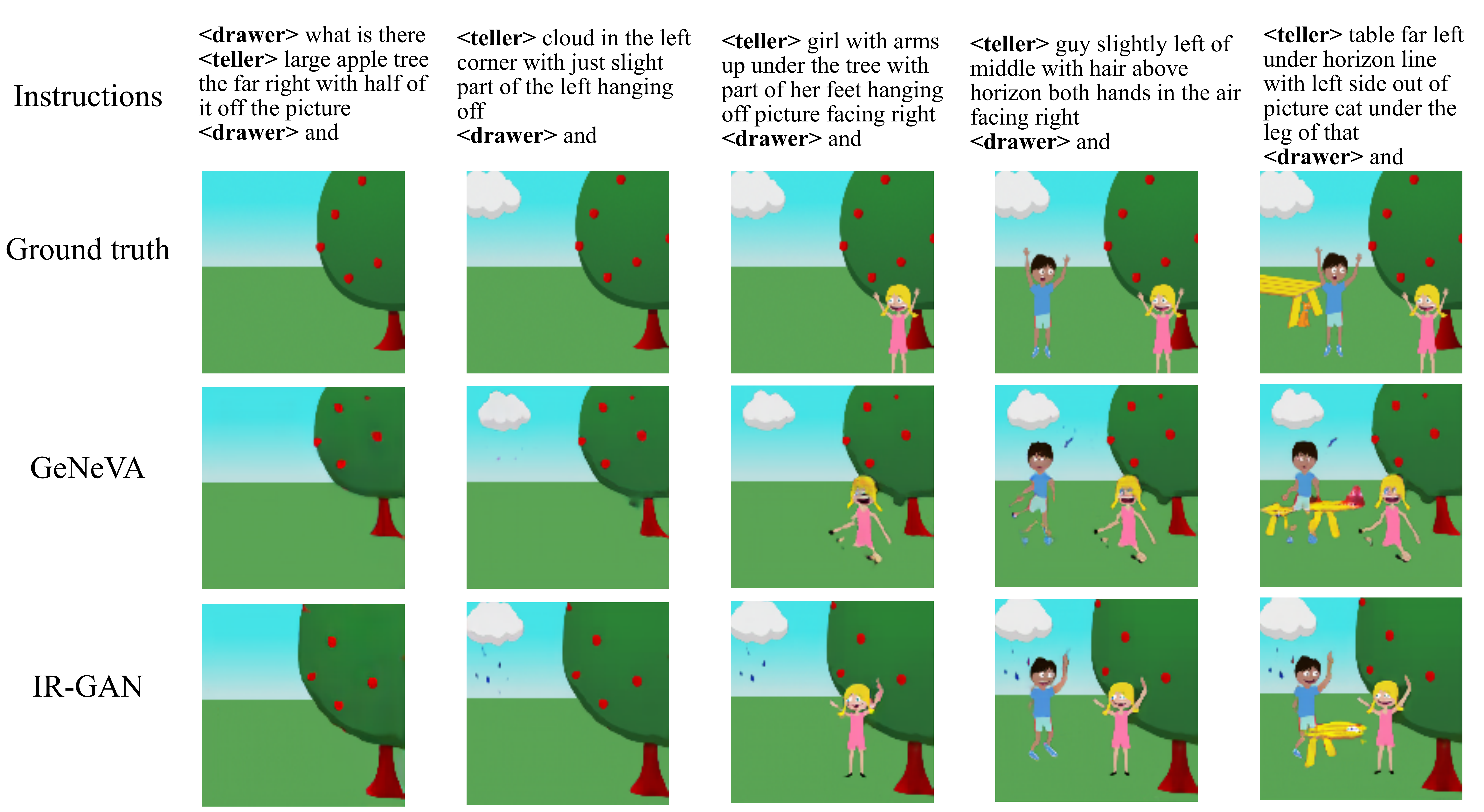}
    \caption{Examples of image manipulation with linguistic instruction on CoDraw dataset, which are
        generated by our IR-GAN and state-of-the-art GeNeVA~\cite{el2019tell}}
    \label{fig:codraw_example}
    \Description{..}
\end{figure*}

\begin{table*}[h]
    \centering
    \caption{Ablation experimental results of different variants conducted on i-CLEVR and CoDraw dataset.}
    \label{tab:ablation}
    \begin{tabular}{|c|l|l|l|l|l|l|l|l|}
        \hline
        \multicolumn{1}{|c|}{Dataset}            & \multicolumn{4}{c|}{i-CLEVR} & \multicolumn{4}{c|}{CoDraw}                                                                                                       \\
        \hline
        Method                                   & Precision                    & Recall                      & F1             & Rsim           & Precision      & Recall         & F1             & Rsim           \\
        \hline
        IR-GAN w/o increment reasoning           & 90.55                        & 82.81                       & 86.51          & 68.75          & 65.12          & 51.10          & 57.26          & 33.40          \\
        IR-GAN w/o historical instruction        & 91.66                        & 82.94                       & 87.08          & 69.52          & 70.10          & 52.72          & 60.18          & 34.45          \\
        IR-GAN w/o user's intention purification & 90.78                        & 84.83                       & 87.70          & 73.08          & 70.00          & 52.09          & 59.73          & 34.72          \\
        IR-GAN                                   & \textbf{94.81}               & \textbf{86.90}              & \textbf{90.68} & \textbf{74.26} & \textbf{70.12} & \textbf{52.75} & \textbf{60.20} & \textbf{35.51} \\
        \hline
    \end{tabular}
\end{table*}

\subsection{Qualitative Results}
We present some examples generated by our IR-GAN and state-of-the-art method GeNeVA~\cite{el2019tell},
Figure~\ref{fig:iclevr_example} shows the qualitative comparison on i-CLEVR dataset and Figure~\ref{fig:codraw_example} shows that on CoDraw dataset.

For the i-CLEVR dataset, we can find, first, our model can better capture the color
and shape of objects in instructions accurately and generate them on the right place of the target image, which proves the ability of
our IR-GAN to understand and reason the spatial information and content from source image and linguistic instruction.
Second, comparing with GeNeVA~\cite{el2019tell}, the visual elements in images generated by our model are more realistic and the layout are more reasonable.
Moreover, observed from the right first column of Figure~\ref{fig:iclevr_example}, our IR-GAN can generate new object, which is partially covered by objects of source image.
Third, comparing with ground truth image, our model has the difficulty in generating logical shaders due to the weak supervision information about light direction.

For the CoDraw dataset, a series of generated target images show that our IR-GAN is able to generate scenes that represent most information in linguistic instruction.
In detail, we can see that:
(1) The elements especially the large elements like trees and cloud are generated accurately;
(2) the relative spatial layout are placed consistent with linguistic instruction, which proves that our IR-GAN can roughly learn the user's intention
(3) detailed visual elements in the ground truth image are hard to represent.
Besides, compared with state-of-the-art GeNeVA~\cite{el2019tell}, the visual elements generated by our IR-GAN are more close to ground truth image and
captures more semantic information in instructions like human pose.
Finally, since our method focuses on the reasoning process in the multimodal conditional generation task,
we do not apply techniques like stacked generators to improve the resolution of the generated image.

\subsection{Ablation Studies}
We conduct the experiments on different variants of our model to evaluate the effectiveness of each component in IR-GAN, including the increment reasoning, history instruction information, user's intention purification.
The ablation study on i-CLEVR and CoDraw dataset is shown in Table~\ref{tab:ablation}.

\subsubsection{Effects of Increment Reasoning}
To evaluate the performance improvement from the increment reasoning mechanism, we implement a degraded model which just uses
concatenation to composite visual and semantic features.
Specifically, for the generator,
the image decoder uses the concatenation of $L_{G_t}$ and $V_{G_t}$ as the input feature to generate the target image.
For the discriminator,  we do the concatenation of $V_{D_t}$ and $V_{D_{t-1}}$ as its input to measure the consistency with instruction representation $h_t$.

As seen in Table~~\ref{tab:ablation}, benefiting from the increment reasoning,
on the i-CLEVR dataset, IR-GAN boosts the F1 score with +4.17\% and the relational similarity (Rsim) with +5.51\%.
On the CoDraw dataset, it improves the F1 score with +2.94\% and the relational similarity (Rsim) with +2.11\%.
Moreover, this module brings the larger improvement performance than other two modules.
The increment reasoning mechanism formulates the interactions between image and instruction, and helps the model reason the consistency between visual increment in target image and user's intention in linguistic instruction.

% Besides, we observe that this degraded model generates many false images, such as some element is removed without corresponding instruction, or the place of some object is wrong, or the background elements in generated image are not consistent with source image.
% This indicates that without increment reasoning,
% the generator is so vulnerable that it ignores some information in the source image.

\subsubsection{Effects of History Instruction Information}
To evaluate the effects of historical instruction information, we conduct experiments on IR-GAN with only word-level instruction encoder.
Specifically, the $h_t$ is replaced by $d_t$ for both generator and discriminator as condition,
which makes the model only learn information about existing elements from source image.

By using historical instruction information in our model, on the i-CLEVR dataset, the precision, recall, F1 score and relational similarity
score (Rsim) achieve the performance improvement with +3.15\%, +3.96\%, +3.60\% and +4.74\% respectively.
On CoDraw dataset, the precision, recall, F1 score, the Rsim are boosted with +0.02\%, +0.03\%, +0.02\%, and +1.06\% respectively.
From the comparison, we can see that (1) the sequential instructions in i-CLEVR dataset are more correlated and IR-GAN can use this correlation to generate high quality images;
(2) in CoDraw dataset, the historical instruction information is less correlated with current manipulation, but
our IR-GAN has to capture more information from source image to cognize the user's intention.
(3) the increased Rsim score (+1.06\%) shows that although historical instructions in CoDraw include few useful information, IR-GAN is still able to reason the layout of target image.

\subsubsection{Effects of User's Intention Purification}
In our IR-GAN, the learned user's intention representation is used in two ways:
(1) as condition to generate target image;
(2) as information for discriminator to measure the logic of generated image.
We only optimize the instruction encoder in the second procedure.
To evaluate the effectiveness of this module, we design a model which uses two instruction encoders to learn separate representations
for the generator and discriminator respectively.

As shown in Table~\ref{tab:ablation}, the comparison result indicates that purifying user's intention through discriminator helps improve the performance of all metrics on both i-CLEVR and CoDraw dataset.
The reasoning process of discriminator encourages the instruction encoder to learn a better representation that captures the manipulation intention for source image. Besides, the optimization of two instruction encoders consumes more memory and computing resources.

In IR-GAN, the instruction encoder is optimized through the backpropagation of discriminator. Here we also conduct the ablation experiment with training the instruction encoder by the backpropagation of generator. We find that, this new training fails converging and obtains a very low performance.

\section{Conclusion}
In this paper, we propose the Increment Reasoning Generative Adversarial Network (IR-GAN) to
address the multimodal conditional generation task, i.e. image manipulation with linguistic instruction.
The instruction encoder is designed to learn user's intention from current and history instruction, a image generator is used to generate target image based on the features of user's intention and source image.
Finally, the increment reasoning discriminator is proposed to reason the consistency between visual increment upon images and semantic increment in linguistic instruction, which ensures the quality and good logic of generated image.
Extensive experiments and ablation studies on CoDraw and i-CLEVR datasets show the effectiveness of our IR-GAN.

\section*{Acknowledgement}
This work was supported in part by the National Key R\&D Program of China under Grant 2018AAA0102003, in part by National Natural Science Foundation of China: 61771457, 61732007, 61620106009, U1636214, 61931008, 61836002, 61672497, in part by Key Research Program of Frontier Sciences, CAS: QYZDJ-SSW-SYS013.
%\newpage

\bibliographystyle{ACM-Reference-Format}
\balance
\bibliography{sample-sigconf}

\end{document}